\documentclass{llncs}
\usepackage{nando-aaai}

\usepackage{url}
\usepackage{picinpar}
\usepackage{floatrow}
\usepackage{xcolor}
\usepackage{fancyvrb, listings}

\usepackage{array, caption, floatrow, tabularx, makecell, booktabs}%
\setcellgapes{3pt}

\urldef{\mailsUS}\path|{wkluegel,miqbal,wyeoh,epontell}@cs.nmsu.edu|
\urldef{\mailsITA}\path|{fioretto}@umich.edu|

\newcommand\tinytt[1]{\textsl{\scriptsize{#1}}}

\title{A Realistic Dataset for the Smart Home Device Scheduling Problem for DCOPs
\vspace{-8pt}}

\author{William Kluegel\inst{1} \and
		Muhammad Aamir Iqbal\inst{1} \and
		Ferdinando Fioretto\inst{2} \and \\
        William Yeoh\inst{1} 
        \and
        Enrico Pontelli\inst{1} 
        \vspace{-4pt}
}
\institute{
Department of Computer Science, New Mexico State University \and
Department of Industrial and Operations Engineering, University of Michigan \\
\mailsUS, \mailsITA
\vspace{-8pt}
}

\begin{document}

\maketitle \sloppy \allowdisplaybreaks

\begin{abstract}
The field of Distributed Constraint Optimization has gained momentum in recent years thanks to its ability to address various applications related to multi-agent cooperation. 
While techniques to solve Distributed Constraint Optimization Problems (DCOPs) are abundant and have matured substantially since the field inception, the number of DCOP realistic applications and benchmark used to asses the performance of DCOP algorithms is lagging behind. 
To contrast this background we 
\emph{(i)} introduce the Smart Home Device Scheduling (SHDS) problem, which describe the problem of coordinating smart devices schedules across multiple homes as a multi-agent system,
\emph{(ii)} detail the physical models adopted to simulate smart sensors, smart actuators, and homes environments, and 
\emph{(iii)} introduce a DCOP realistic benchmark for SHDS problems. 
\end{abstract}

\section{Introduction}

\emph{Distributed Constraint Optimization Problems (DCOPs)} \cite{modi:05,petcu:05,yeoh:12} have emerged as one of the prominent agent models to govern the agents' autonomous behavior, where both algorithms and communication models are driven by the structure of the specific problem. Since the research field inception 
a wide variety of algorithms have been proposed to solve DCOPs and typically classified as being either \emph{complete} or \emph{incomplete}, based on whether they can guarantee the optimal solution or they trade optimality for shorter execution times. In addition, each of these classes can be categorized into several groups, depending on the degree of locality exploited by the algorithms (e.g., partial centralization) \cite{hirayama:97,mailler:04,petcu:07b}, the way local information is updated (e.g., synchronous \cite{mailler:04,pearce:07,petcu:05} or asynchronous \cite{farinelli:08,gershman:09,modi:05}), and the type of exploration process adopted (e.g., search-based \cite{hirayama:97,modi:05,zhang:05}, inference-based \cite{petcu:05,farinelli:08}, or sampling-based \cite{ottens:12,nguyen:13,fioretto:cp16}). 

While techniques to solve DCOPs are abundant and have matured substantially since the field inception, the number of DCOP realistic applications and benchmarks used to assess the performance of DCOP algorithms is lagging behind. Typical DCOP algorithms are evaluated on artificial random problems, or simplified problems that are adapted to the often unrealistic assumptions made by DCOP algorithms (e.g., that each agent controls exactly one variable, and that all problem constraints are binary). 
To assess the performance of DCOP algorithms it is necessary to introduce realistic problem benchmark of deployable applications.

Motivated by these issues, we recently introduced the \emph{Smart Home Device Scheduling (SHDS)} problem \cite{fioretto:aamas-17b}, which formalizes the problem of coordinating smart devices (e.g.,~smart thermostats, circulator heating, washing machines) schedules across multiple smart homes as a multi-agent system (MAS). The SHDS problem is suitable to be modeled as a DCOP due to the presence of both complex individual agents' goals, describing homes' energy price consumption, as well as a collective agents' goal, capturing the energy peaks reduction. 

In this document we introduce a set of realistic synthetic benchmarks for the SHDS problem for DCOPs. We report the details of the physical models adopted to simulate smart home sensors and actuators, as well as home environments, and describe how the actuator's actions affects the environments of a home (e.g., home's temperature, cleanliness, humidity). 
The datasets, models, and code adopted to generate the SHDS datasets are available at: \url{https://github.com/persoon/SHDS_dataset}. 


\vspace{-10pt}
\subsubsection{DCOP}

A \emph{Distributed Constraint Optimization Problem (DCOP)}~\cite{modi:05,yeoh:12} is described by a tuple $\langle\mathcal{X, D, F, A, \,}\alpha\rangle$, 
where:
$\mathcal{X} \!=\! \{x_1,\ldots,x_n\}$ is a set of \emph{variables}; 
$\mathcal{D} \!=\! \{D_1,\ldots,D_n\}$ is a set of finite \emph{domains} (i.e.,~$x_i \!\in\!  D_i$); 
$\mathcal{F} \!=\! \{f_1, \ldots, f_e\}$ is a set of \emph{utility functions} (also called \emph{constraints}), where $f_i: \bigtimes_{x_j \in \scope{f_i}} D_i \rightarrow \mathbb{R}_+ \cup \{-\infty\}$ and 
$\scope{f_i} \!\subseteq\! \mathcal{X}$ is the set of the variables (also called the \emph{scope}) relevant to $f_i$;
$\mathcal{A} \!=\! \{a_1, \ldots, a_p\}$ is a set of \emph{agents};
and $\alpha: \mathcal{X} \rightarrow \mathcal{A}$ is a function that maps each variable to one agent. 
$f_i$ specifies the utility of each combination of values assigned to the variables in $\scope{f_i}$.
A \emph{partial assignment} $\sigma$ is a value assignment to a set of variables  $X_{\sigma} \!\subseteq\! \mathcal{X}$  that is consistent with the 
variables' domains. The utility $\mathcal{F}(\sigma) \!=\! \sum_{f\in{\cal F}, \scope{f} \subseteq X_{\sigma}} f(\sigma)$ 
is the sum of the utilities of all the applicable utility functions in $\sigma$. 
A \emph{solution} is a partial assignment $\sigma$ for all the variables of the problem, i.e., with $X_{\sigma} \is \mathcal X$. 
We will denote with $\setf{x}$ a solution, while $\setf{x}_i$ is the value of $x_i$ in $\setf{x}$. The goal is to find an optimal solution $\setf{x}^* = \argmax_{\setf{x}} \mathcal{F}(\setf{x})$.

\section{Scheduling Device in Smart Homes}
\vspace{-4pt}

\begin{figure}[t]
\centering\includegraphics[width=\textwidth]{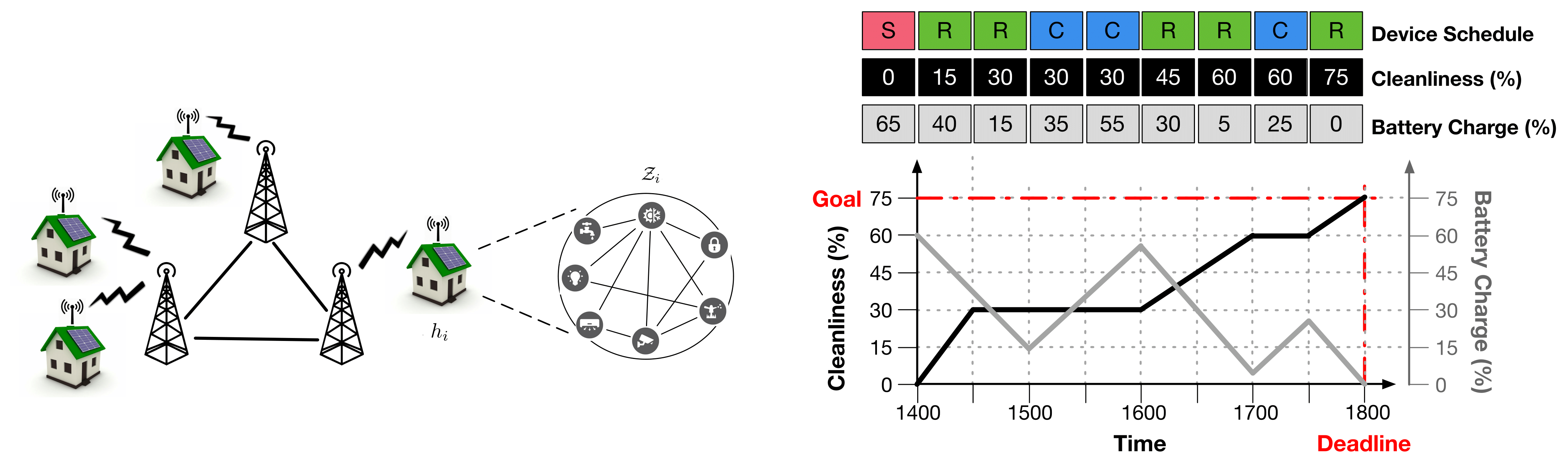}
\caption{Illustration of a Neighborhood of Smart Homes \label{fig:neighborhood}}
\end{figure}

\noindent A \emph{Smart Home Device Scheduling (SHDS)} problem is defined by the tuple $\langle \setf{H}, \mathcal{Z}, \mathcal{L}, \setf{P}_H, \setf{P}_Z, H, \theta \rangle$, where:
$\setf{H} = \{h_1, h_2, \ldots \}$ is a neighborhood of smart homes, capable of communicating  with one another; 
$\mathcal{Z} = \cup_{h_i \in \setf{H}} \setf{Z}_i$ is a set of smart devices, where $\setf{Z}_i$ is the set of devices in the smart home $h_i$ (e.g.,~vacuum cleaning robot, smart thermostat).
$\mathcal{L} = \cup_{h_i \in \setf{H}} \setf{L}_i$ is a set of locations, where $\setf{L}_i$ is the set of locations in the smart home $h_i$ (e.g.,~living room, kitchen);
$\setf{P}_H$ is the set of state properties of the smart homes (e.g.,~cleanliness, temperature); 
$\setf{P}_Z$ is the set of devices state properties (e.g.,~battery charge for a vacuum robot); 
$H$ is the planning horizon of the problem. We denote with $\setf{T}= \{1, \ldots, H\}$ the set of time points; 
$\theta : \setf{T} \to \mathbb{R}^+$ represents the real-time pricing schema adopted by the energy utility company, which expresses the cost per kWh of energy consumed by consumers. 
Finally, we use $\Omega_p$ to denote the set of all possible states for state property $p \in \setf{P}_H \cup \setf{P}_Z$ (e.g.,~all the different levels of cleanliness for the cleanliness property). Figure~\ref{fig:neighborhood}(right) shows an illustration of a neighborhood of smart homes with each home controlling a set of smart devices.

\subsection{Smart Devices}
\label{sec:smart_devices}
\vspace{-4pt}

\noindent For each home $h_i \in \setf{H}$, the set of smart devices $\setf{Z}_i$ is partitioned into a set of actuators 
$\setf{A}_i$ and a set of sensors $\setf{S}_i$. 
Actuators can affect the states of the home (e.g.,~heaters and ovens can affect the temperature in the home) and possibly their own states (e.g.,~ vacuum cleaning robots drain their battery power when running). On the other hand, sensors monitor the states of the home.
Each device $z \in \setf{Z}_i$ of a home $h_i$ is defined by a tuple $\langle \ell_z, A_z, \gamma_z^H, \gamma_z^Z \rangle$, where $\ell_z \in \setf{L}_i$ denotes the relevant location in the home that it can act or sense, $A_z$ is the set of actions that it can perform, $\gamma_z^H: A_z \to 2^{\setf{P}_H}$ maps the actions of the device to the relevant state properties of the home, and $\gamma_z^Z: A_z \to 2^{\setf{P}_Z}$ maps the actions of the device to its relevant state properties. We will use the following running example throughout this paper. 

\begin{example}
Consider a vacuum cleaning robot $z_v$ with location $\ell_{z_v} \!=\! \statef{living\_room}$.
The set of possible actions is $A_{z_v} = \{\statef{run}, \statef{charge}, \statef{stop}\}$ and the mappings are:
\begin{align*}
\footnotesize
\gamma_{z_v}^H &\!\!: \statef{run} \!\to\! \{\text{cleanliness}\}; 
\hspace{0.1em} 		  \statef{charge} \!\to\! \emptyset;
\hspace{0.1em} 		  \statef{stop} \!\to\! \emptyset \\
\gamma_{z_v}^Z &\!\!: \statef{run} \!\to\! \{\text{battery\_charge}\}; 
\hspace{0.1em} 		  \statef{charge} \!\to\! \{\text{battery\_charge}\}; \statef{stop} \!\to\! \emptyset 
\end{align*}
where $\emptyset$ represents a \emph{null} state property.
\end{example}

\subsection{Device Schedules} 
\label{sec:dev_schedules}
\vspace{-4pt}

\noindent To control the energy profile of a smart home we need to describe the behavior of the smart devices acting in the smart home during time. We formalize this concept with the notion of \emph{device schedules}.

We use $\xi_{z}^t \in A_z$ to denote the action of device $z$ at time step $t$, and $\xi_{X}^t = \{\xi_{z}^t \st z \in X\}$ to denote the set of actions of the devices in $X \subseteq \mathcal{Z}$ at time step $t$.

\begin{definition}[Schedule]$\!\!\!$
A schedule $\xi_{X}^{[t_a \!\to t_b\!]} \!=\! \langle \xi_{X}^{t_a}, \ldots, \xi_{X}^{t_b} \rangle$ is a sequence of actions for the devices in $X \subseteq \mathcal{Z}$ within the time interval from $t_a$ to $t_b$.
\end{definition}


Consider the illustration of Figure~\ref{fig:neighborhood}(left). The top row of Figure~\ref{fig:neighborhood}(left) shows a possible schedule $\langle R, R, C, C, R, R, C, R \rangle$ for a vacuum cleaning robot starting at time 1400 hrs, where each time step is 30 minutes. The robot's actions at each time step are shown in the colored boxes with letters in them: red with `S' for \statef{stop}, green with `R' for \statef{run}, and blue with `C' for \statef{charge}. 

At a high-level, the goal of the SHDS problem is to find a schedule for each of the devices in every smart home that achieve some user-defined objectives (e.g.,~the home is at a particular temperature within a time window, the home is at a certain cleanliness level by some deadline) that may be personalized for each home. We refer to these objectives as \emph{scheduling rules}.

\subsection{Scheduling Rules}
\vspace{-4pt}

\noindent We define two types of scheduling rules: 
\emph{Active scheduling rules (ASRs)} that define user-defined objectives on a desired state of the home (e.g.,~the living room is cleaned by 1800 hrs), and 
\emph{Passive scheduling rules (PSRs)} that define implicit constraints on devices that must hold at all times (e.g., the battery charge on a vacuum cleaning robot is always between 0\% and 100\%). 
We provide a formal description for the grammar of scheduling rules in Section \ref{sec:schduling_rules}.

\begin{example}
\label{ex:2}
The scheduling rule \eqref{active_rule} describes an \emph{ASR} defining a goal state where the living room floor is at least 75\% clean (i.e.,~at least 75\% of the floor is cleaned by a vacuum cleaning robot) by 1800 hrs:
\begin{align}
&\texttt{\footnotesize living\_room cleanliness $\geq$ 75\! before\! 1800}  \label{active_rule} \\
&\texttt{\footnotesize $z_v$ battery\_charge $\geq$ 0 always}  \label{passive_rule} \\
&\texttt{\footnotesize $z_v$ battery\_charge $\leq$ 100 always} \label{passive_rule2}
\end{align}
and scheduling rules \eqref{passive_rule} and \eqref{passive_rule2} describe \emph{PSRs} stating that the battery charge of the vacuum robot $z_v$ needs to be between 0 and 100 \% of its full charge at all the times:
\end{example}

%
We denote with $R_{p}^{[t_a \!\to t_b]}\!$ a scheduling rule over a state property $p \!\in\! \setf{P}_H \!\cup\! \setf{P}_Z$, and time interval $[t_a, t_b]$. 
Each scheduling rule indicates a goal state at a location or on a device $\ell_{R_p} \!\in\! \setf{L}_i \!\cup\! \setf{Z}_i$ of a particular state property $p$ that must hold over the time interval $[t_a, t_b] \subseteq \setf{T}$. 
The scheduling rule goal state is either a desired state of a home, if it is an ASR (e.g., the cleanliness level of the room floor) or a required state of a device or a home, if it is a PSR (e.g., the battery charge of the vacuum cleaning robot).

Each rule is associated with a set of actuators $\Phi_p \subseteq \setf{A}_i$ that can be used to reach the goal state. For instance, in our Example (\ref{ex:2}), $\Phi_p$ correspond to the vacuum cleaning robot $z_v$, which can operate on the living room floor. 
Additionally, a rule is associated with a sensor $s_p \!\in\! \setf{S}_i$ capable of sensing the state property $p$. Finally, in a PSRs the device can also sense its own internal states. 

The ASR of Equation~\eqref{active_rule} is illustrated in Figure~\ref{fig:neighborhood}(left) by dotted red lines on the graph. The PSRs are not shown as they must hold for all time steps.

\subsection{Feasibility of Schedules}
\label{sec:feas_schedule}
\vspace{-4pt}

\noindent To ensure that a goal state can be achieved across the desired time window the system uses a \emph{predictive model} of the various state properties. This predictive model captures the evolution of a state property over time and how such state property is affected by a given joint action of the relevant actuators. We describe the details of the physical predictive models used to generate our benchmark set in Section \ref{sec:ds_pred_models}.

\begin{definition}[Predictive Model]
A predictive model $\Gamma_{p}$ for a state property $p$ (of either the home or a device) is a function
$\Gamma_{p}  : \Omega_p \times \bigtimes_{z \in \Phi_p} A_z \cup \{\bot\}\to \Omega_p \cup \{\bot\}$, where $\bot$ denotes an infeasible state and $\bot + (\cdot) = \bot$.
\end{definition}

In other words, the model describes the transition of state property $p$ from state $\omega_p \in \Omega_p$ at time step $t$ to time step $t+1$ when it is affected by a set of actuators $\Phi_p$ running joint actions $\xi_{\Phi_p}^t$:
\begin{align}	\label{eq:sensorModel}
	\Gamma_{p}^{t+1}(\omega_p, \xi_{\Phi_p}^t) =
		\omega_p + \Delta_{p} (\omega_p, \xi_{\Phi_p}^t)
\end{align}
where $\Delta_{p}(\omega_p, \xi_{\Phi_p}^t)$ is a function describing the effect of the actuators' joint action $\xi_{\Phi_p}^t$ on state property $p$. 
We assume here, w.l.o.g. that the state of properties are numeric---when this is not the case, a mapping to the possible states to a numeric representation can be easily defined.

Notice that a recursive invocation of a predictive model allows us 
to predict the trajectory of a state property $p$ for future time steps, given a schedule of actions of the relevant actuators $\Phi_p$. 
Let us formally define this concept.

\begin{definition}[Predicted State Trajectory]
Given a state property $p$, its current state $\omega_p$ at time step $t_a$, and a schedule $\xi_{\Phi_p}^{[t_a \to t_b]}$ of relevant actuators $\Phi_p$, the predicted state trajectory $\pi_{p} (\omega_p, \xi_{\Phi_p}^{[t_a \to t_b]})$ of that state property is defined as:
\begin{align}
\pi_{p} (\omega_p, \xi_{\Phi_p}^{[t_a \to t_b]}) &= 
	\Gamma_{p}^{t_b}( \Gamma_{p}^{t_{b-1}}( \ldots (
		\Gamma_{p}^{t_a}(\omega_p, \xi_{\Phi_p}^{t_a}),
		\ldots), \xi_{\Phi_p}^{t_{b-1}}), \xi_{\Phi_p}^{t_b}) 
\end{align}
\end{definition}

Consider the device scheduling example in Figure \ref{fig:neighborhood}(left). The predicted state trajectories of the \emph{battery\_charge} and \emph{cleanliness} state properties 
are shown in the second and third rows of Figure \ref{fig:neighborhood}(left). These trajectories are predicted given that the vacuum cleaning robot will take on the schedule shown in the first row of the figure. The predicted trajectories of these state properties are also illustrated in the graph, where the dark grey line shows the states for the robot's battery charge and the black line shows the states for the cleanliness of the room. 

Notice that to verify if a schedule satisfies a scheduling rule 
it is sufficient to check that the predicted state trajectories are within the set of feasible state trajectories of that rule.
Additionally, notice that each active and passive scheduling rule defines a set of feasible state trajectories. For example, the active scheduling rule of Equation~\eqref{active_rule} allows all possible state trajectories as long as the state at time step 1800 is no smaller than 75. 
We use $R_p[t] \subseteq \Omega_p$ to denote the set of states that are feasible according to rule $R_p$ of state property $p$ at time step $t$.
More formally, a schedule $\xi_{\Phi_p}^{[t_a \to t_b]}$ satisfies a scheduling rule $R_p^{[t_a \to t_b]}$ (written as $\xi_{\Phi_p}^{[t_a \to t_b]} \models R_p^{[t_a \to t_b]}$) iff:
\begin{align}
	\forall t \in [t_a, t_b]: 
	\pi_{p}(\omega_p^{t_a},  \xi_{\Phi_p}^{[t_a \to t]})
	\in R_{p}[t]
\end{align}
where $\omega_p^{t_a}$ is the state of state property $p$ at time step $t_a$.
\begin{definition}[Feasible Schedule]
A schedule is \emph{feasible} if it satisfies \emph{all} the passive and active scheduling rules of each home in the SHDS problem.
\end{definition}

In the example of Figure \ref{fig:neighborhood}, the evaluated schedule is a feasible schedule since the trajectories of both the \emph{battery\_charge} and \emph{cleanliness} states satisfy both the \emph{active scheduling rule} \eqref{active_rule} and the \emph{passive scheduling rules} \eqref{passive_rule} and~\eqref{passive_rule2}.

\subsection{Optimization Objective}
\vspace{-4pt}

\noindent In addition to finding feasible schedules, the goal in the SHDS problem is to optimize for the aggregated total cost of energy consumed.

Each action $a \!\in\! A_z$ of device $z \!\in\! \setf{Z}_i$ in home $h_i \!\in\! \setf{H}$ has an associated energy consumption $\rho_z \!: A_z \!\to\! \mathbb{R}^+$, expressed in kWh. The aggregated energy $E_i^t(\xi_{\setf{Z}_i}^{[0 \to H]})$ across all devices consumed by $h_i$ at time step $t$ under trajectory $\xi_{\setf{Z}_i}^{[0 \to H]}$ is:
\begin{equation}
E_i^t(\xi_{\setf{Z}_i}^{[0 \to H]}) = \sum_{z \in \setf{Z}_i}  \rho_{z}(\xi_z^t)
\end{equation} 
where $\xi_z^t$ is the action of device $z$ at time $t$ in the schedule $\xi_{\setf{Z}_i}^{[0 \to H]}$. 
The cost $c_i (\xi_{\setf{Z}_i}^{[0 \to H]})$ associated to schedule $\xi_{\setf{Z}_i}^{[0 \to H]}$ in home $h_i$ is:
\begin{equation} 	\label{eq:cost}
	c_i (\xi_{\setf{Z}_i}^{[0 \to H]}) = \sum_{t \in \setf{T}} 
			\big(\ell_i^t  + E_i^t(\xi_{\setf{Z}_i}^{[0 \to H]}) ) \cdot \theta(t)
\end{equation}
where $\ell_i^t$ is the home background load produced at time $t$, which includes all non-schedulable devices (e.g.,~TV, refrigerator), and sensor devices, which are always active, and $\theta(t)$ is the real-time price of energy per kWh at time $t$.

\smallskip
The objective of an SHDS problem is that of minimizing the following weighted bi-objective function:
\begin{align}
& \hspace{40pt} \min_{\xi_{\setf{Z}_i}^{[0 \to H]}} \quad \alpha_c \!\cdot\!C^{\text{sum}} + \alpha_e \!\cdot\!E^{\text{peak}}  \label{eq:obj} \\
 \hspace{-10pt} \text{subject to:}  \qquad& 
	\forall h_i \in \setf{H}, R_p^{[t_a \to t_b]} \in \setf{R}_i: \quad \xi_{\Phi_p}^{[t_a \to t_b]} \models R_p^{[t_a \to t_b]} 
	\label{eq:constr_2}
\end{align} 
where $\alpha_c, \alpha_e \!\in\! \mathbb{R}$ are weights, 
$
C^{\text{sum}} = \sum_{h_i \in \setf{H}} c_i(\xi_{\setf{Z}_i}^{[0 \to H]})
$ 
is the aggregated monetary cost across all homes $h_i$; and
$
E^{\text{peak}} = \sum_{t \in \setf{T}} \
	\sum_{\setf{H}_j \in \mathcal{H}} 
	\sum_{h_i \in \setf{H}_j} 
	\big(E_i^t(\xi_{\setf{Z}_i}^{[0 \to H]}) \big)^2
$ 
is a quadratic penalty function on the aggregated energy consumption across all homes $h_i$. 
Since the SHDS problem is designed for distributed multi-agent systems, in a cooperative approach optimizing $E^{\text{peak}}$ may require each home to share its energy profile with each other home. To take into account data privacy concerns and possible high network loads, we decompose the set of homes $\setf{H}$ into neighboring subsets of homes $\mathcal{H}$, so that $E^{\text{peak}}$ can be optimized independently within each subset. 
 These coalitions can be exploited by a distributed algorithm to (1) parallelize computations between multiple groups and (2) avoid data exposure over long distances or sensitive areas. 
Finally, constraint~\eqref{eq:constr_2} defines the valid trajectories for each scheduling rule $r \in \setf{R}_i$, where $\setf{R}_i$ is the set of all scheduling rules of home $h_i$.

\subsection{DCOP Mapping}
\vspace{-4pt}

One can map the SHDS problem to a DCOP as follows:
\bitemize 
\item \textsc{Agents:} Each agent $a_i \in \mathcal{A}$ in the DCOP is mapped to a home $h_i \in \setf{H}$.

\item \textsc{Variables} and \textsc{Domains:} Each agent $a_i$ controls the following set of variables:
\bitemize
\item For each actuator $z \in \setf{A}_i$ and each time step $t \in \setf{T}$, a variable $x_{i,z}^t$ whose domain is the set of actions in $A_z$. 
The sensors in $\setf{S}_i$ are considered to be always active, and thus not directly controlled by the agent. 
\item An auxiliary interface variable $\hat{x}_j^t$ whose domain is the set $\{0, \ldots, \sum_{z \in \setf{Z}_i } \rho(\argmax_{a \in A_z}  \rho_z(a)) \}$, which represents the aggregated energy consumed by all the devices in the home at each time step $t$.
\eitemize

\item \textsc{Constraints:} There are three types of constraints:
\bitemize
\item \emph{Local} soft constraints (i.e.,~constraints that involve only variables controlled by the agent) whose costs correspond to the weighted summation of monetary costs, as defined in Equation~\eqref{eq:cost}. 
\item \emph{Local} hard constraints that enforce Constraint~\eqref{eq:constr_2}. Feasible schedules incur a cost of 0 while infeasible schedules incur a cost of $\infty$.
\item \emph{Global} soft constraints (i.e.,~constraints that involve variables controlled by different agents) whose costs correspond to the peak energy consumption, as defined in the second term in Equation~\eqref{eq:obj}. 
\eitemize
\eitemize

\section{Model Parameters and Realistic Data Set Generation}
\label{sec:model_params}
\vspace{-4pt}

This section describes the parameters and models adopted in our SHDS datasets generation. 
We first describe the house structural parameters, which are used in turn to calculate the house predictive models. Next, we report a detailed list of the smart devices adopted in our datasets, discussing their power consumptions and effects on the house environments. We then describe the predictive models adopted to capture changes in the house's environments and devices' states. Finally, we report the BNF for the scheduling rules introduced in Section 2.3, and the pricing scheme adopted in our experiments.

\subsection{House Structural Parameters}
\label{sec:house_params}
\vspace{-4pt}

\begin{figure}[!tb]
\centering
\includegraphics[height=1.3in]{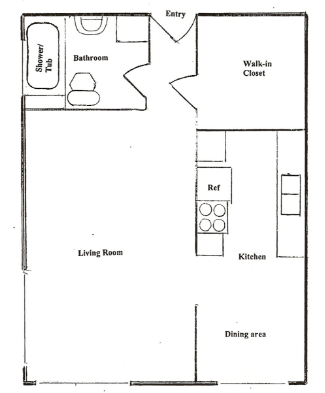}
\includegraphics[height=1.3in]{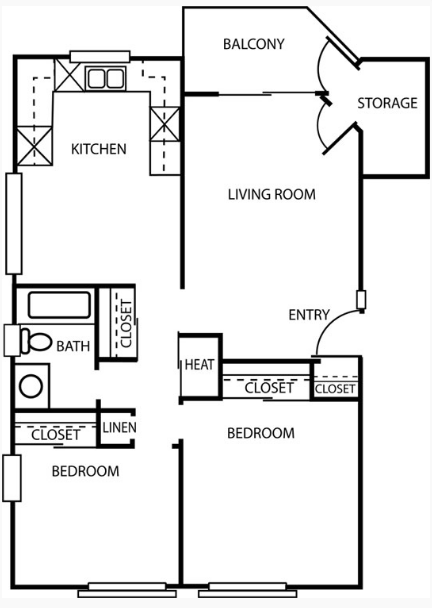}
\includegraphics[height=1.3in]{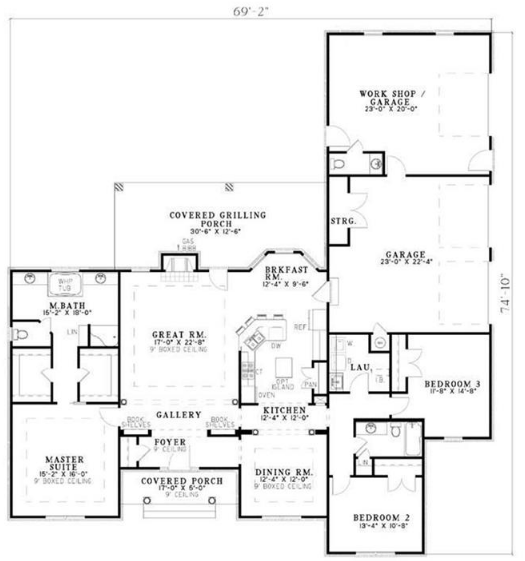}
\caption{\label{tab:structure} Floor plans for a small (left), medium (center), and large (right) house.}
\end{figure}

We consider three house sizes (small, medium, and large). 
The floor plans for three house structures are shown in Figure \ref{tab:structure}. 
Our house structural model simplifies the floor plans shown in Figure \ref{tab:structure} by ignoring internal walls. 
This abstraction is sufficient to capture the richness of the predictive models introduced in Section \ref{sec:feas_schedule}.
Table \ref{tab:house} reports the parameters of the houses adopted in our SHDS dataset. 
The house sizes are expressed in meters ($L \times W$). 
The walls height is assumed to be $2.4\, m$ and the window area denotes the area of the walls covered by windows. 
The overall heat transfer coefficient (also referred to as U-value) describes how well a building element conducts 
heat. It is defined as the rate of heat transfer (in watts) through one unit area ($m^2$) of a structure divided 
by the difference in temperature across the structure \cite{mitchell2012principles}. 

\memo{Fix Reference here}

The walls material is considered to be a $150\, mm$ poured concrete ($1280\, kg/m^3$) with a Heat-Transfer Coefficient 
($U_\text{walls}$) of $3.9\, \frac{W}{m^2 \cdot~^\circ\,C}$.
We consider vertical double glazed windows, with distance between glasses $30 - 60\, mm$ whose Heat-Transfer Coefficient 
($U_\text{windows}$) is $2.8\, \frac{W}{m^2 \cdot~^\circ\,C}$. 
Additionally, we consider a $2.54\, cm$ wood roof with $2.54\, cm$ insulation, with Heat-Transfer Coefficient 
($U_\text{roof}$) of $1.1\, \frac{W}{m^2 \cdot~^\circ\,C}$. 
Finally, we consider a $5.08\, cm$ wood door, with Heat-Transfer Coefficient of $2.6\, \frac{W}{m^2 \cdot~^\circ\,C}$. 
These are commonly adopted materials in the US house construction industry \cite{mitchell2012principles}.
We assume a background load consumption which account of a medium-size refrigerator ($120\, W$), a wireless router ($6\, W$), and a set of light bulbs (collectively $40\, W$) \cite{mitchell2012principles}. 
The heat gain from the background house appliances is computed according to \cite{mitchell2012principles}(Table 9.8).
We consider the heat gain from people within the house, and computed as in \cite{mitchell2012principles}(Table 9.7), assuming a metabolic rate as \emph{light office work}.

\begin{table}[!t]
\caption{\label{tab:house} House structural parameters.}
\resizebox{0.9\textwidth}{!}
{
\begin{tabular}{l r r r c | l r r r}
\hline
	Structural Parameters & 	small 		& 	medium		& 	large 	
	&~~~~~&
	Structural Parameters & 	small 		& 	medium		& 	large 	\\
	\hline
	house size ($m$) 	 &  	$6 \times 8$	& $8 \times 12$	& $12 \times 15$ 
	& &
	$U_\text{roof}$ $(W/m^2C)$	 & 1.1 & 1.1 & 1.1 \\
	walls area ($m^2$)   &	67.2          &     96           & 129.6 
	& &
	lights energy density ($W/m^3$) &    9.69 & 9.69 & 9.69 \\
	window area ($m^2$)  &	7.2			&	 10			& 16    
	&&
	background load (kW) & 0.166	& 0.166	& 0.166 \\
	$U_\text{walls}$ $(W/m^2C)$	 & 3.9 & 3.9 & 3.9 
	&&
	background heat gain (W)  & 50	& 50		& 50 \\
	$U_\text{windows}$ $(W/m^2C)$ & 2.8 & 2.8 & 2.8 
	&&
	people heat gain  (Btu/h) & 400  & 400    & 400 \\ 
	\hline
	\end{tabular}
}
\end{table}

\subsection{Smart Devices}
\vspace{-4pt}

In this section we report the complete list of smart devices (sensors and actuators) adopted by the smart homes in our SHDS datasets.
 
\smallskip\noindent\textbf{Sensors} 
%
Table \ref{tab:sensors} reports the sensors adopted in our SHDS problem. For each sensor, 
we report an identifier (ID), the state property (see Section \ref{sec:smart_devices}) it senses, 
and its location in the house. All sensors are considered to be constantly active, 
sensing a single state property at a location (e.g., an air temperature sensor is located in a 
house room, a charge sensor is located on a device).  


\begin{table}[!tb]
\caption{List of sensors. \label{tab:sensors}}
\resizebox{0.7\textwidth}{!}
{
	\begin{tabular}{c c c c c c c}
	\hline
	ID	& State property & Location &~~~~~& ID	& State property & Location \\
	\hline
	01		& air temperature 			& house room && 	08		& dish cleanliness     		& appliance  \\
	02		& floor cleanliness (dust)	& house room &&		09		& air humidity	  			& house room \\
	03		& temperature				& appliance  &&		10		& luminosity  				& house room \\
	04		& battery charge			& appliance  &&		11		& occupancy   				& house room \\
	05		& bake						& appliance  &&		12		& movement 					& house room \\
	06		& laundry wash				& appliance  &&		13		& smoke detector			& house room \\
	07		& laundry dry				& appliance  &&		& & \\
\hline 
\end{tabular}
}
\end{table}

\smallskip\noindent\textbf{Actuators}
Table \ref{tab:actuators} reports the list of the actuators.  
It tabulates the type of actuator and its model, its possible actions, 
the power consumption (in kWh), the state properties affected by each of its action, and 
the effects ($\Delta$) on the associated predictive models in the small, medium, and large house sizes. 
The latter represent the incremental quantity which affects the physical system, given the 
action of the actuator, as defined in Equation \ref{eq:sensorModel}. We detail the 
calculation of the house and devices physical models below.

\begin{table}[!t]
\caption{List of actuators. \label{tab:actuators}}
\resizebox{\textwidth}{!}
{
\begin{tabular}{c c l c c c c c}
\hline
Actuator	& Model	& Actions & Consumption (kWh)		& State properties (ID) 	& Effects Small($\Delta$) 	& Effects Medium($\Delta$) 	& Effects Large($\Delta$)\\
\hline
\multirow{3}{*}{Heater} 
	& \multirow{3}{*}{Bryant 697CN030B}
		&  off
			& $0$	& $\{01\}$ 
			& $- \frac{\dot{L}_h}{148.48 \cdot T_A}$ 
			& $- \frac{\dot{L}_h}{296.86 \cdot T_A}$ 
			& $- \frac{\dot{L}_h}{593.75 \cdot T_A}$ \\
		&& fan  
			& $0.5$	& $\{01\}$ 
			& $- \frac{\dot{L}_h}{148.48 \cdot T_A}$ 
			& $- \frac{\dot{L}_h}{296.86 \cdot T_A}$ 
			& $- \frac{\dot{L}_h}{593.75 \cdot T_A}$ \\
		&& heat 
		 	& $0.5$ &  $\{01\}$
			& $\frac{\dot{L}_h}{148.48 \cdot |T_Z - T_A| }$  
			& $\frac{\dot{L}_h}{296.86 \cdot |T_Z - T_A| }$
			& $\frac{\dot{L}_h}{593.75 \cdot |T_Z - T_A| }$ \\
\hline
\multirow{3}{*}{Cooler}
	& \multirow{3}{*}{LG LW1212ER}
		&	off  	
			& $0$	& $\{01\}$
			& $ \frac{\dot{L}_h}{148.48 \cdot T_A}$
			& $ \frac{\dot{L}_h}{296.86 \cdot T_A}$
			& $ \frac{\dot{L}_h}{593.75 \cdot T_A}$ \\
		&& fan
			& $0.5$	& $\{01\}$
			& $ \frac{\dot{L}_h}{148.48 \cdot T_A}$
			& $ \frac{\dot{L}_h}{296.86 \cdot T_A}$
			& $ \frac{\dot{L}_h}{593.75 \cdot T_A}$ \\
		&& cool
			& $2.3$	& $\{01\}$
			& $\frac{\dot{L}_h}{148.48 \cdot |T_A - T_Z| }$  
			& $\frac{\dot{L}_h}{296.86 \cdot |T_A - T_Z| }$
			& $\frac{\dot{L}_h}{593.75 \cdot |T_A - T_Z| }$ \\
\hline
\multirow{2}{*}{Waterheater}  & \multirow{2}{*}{E52-50R-045DV}  &    
			   off  	& $0$ & $\{03\}$  & $\{0\}$ & $\{0\}$ & $\{0\}$ \\
			&& on  		& $4.6$ & $\{03\}$ & $\{27.9^{\circ}C\}$ & $\{25.2^{\circ}C\}$ & $\{19.2^{\circ}C\}$ \\

\hline
\multirow{3}{*}{Vacuum Bot}   & \multirow{3}{*}{iRobot Roomba 880}  &    
			   off  	& $0$ & $\{02,\ 04\}$ & $\{0.0 \% ,\   0.0 \% \}$	& $\{0.0 \% ,\   0.0 \% \}$ & $\{0.0 \% ,\   0.0 \% \}$ \\
			&& vacuum  	& $0$ & $\{02,\ 04\}$ & $\{0.676 \% ,\ -0.21 \% \}$	& $\{0.338 \% ,\ -0.21 \% \}$ & $\{0.168 \% ,\ -0.21 \% \}$ \\
			&& charge  	& $0.26$ & $\{04\}$ & $\{0.33 \% \}$				& $\{0.33 \% \}$ & $\{0.33 \% \}$ \\
\hline
\multirow{2}{*}{Electric Vehicle}		& \multirow{2}{*}{Tesla Model S}  &    
  		  	   off  & $0$ & $\{04\}$ & $\{0\}$ & $\{0\}$ & $\{0\}$ \\
		     && charge & $17.28$ & $\{04\}$ & $\{0.167 \% \}$ & $\{0.167 \% \}$ & $\{0.167 \% \}$ \\
\hline
\multirow{2}{*}{Clothes Washer} & \multirow{2}{*}{GE WSM2420D3WW}  &
			   off   & $0$ & $\{06\}$ & $\{0\}$  & $\{0\}$ & $\{0\}$ \\
			  && wash (Regular) & $0.43$ & $\{06\}$ & $\{1\}$  & $\{1\}$  & $\{1\}$ \\
			  && spin (Regular) & $0.45$ & $\{06\}$   & $\{1\}$  & $\{1\}$ & $\{1\}$ \\
			  && rinse (Regular) & $0.15$ & $\{06\}$  & $\{1\}$  & $\{1\}$ & $\{1\}$ \\

			  && wash (Perm-Press) & $0.39$ & $\{06\}$ & $\{1\}$  & $\{1\}$ & $\{1\}$ \\
			  && spin (Perm-Press) & $0.43$ & $\{06\}$ & $\{1\}$  & $\{1\}$ & $\{1\}$ \\
			  && rinse (Perm-Press) & $0.23$ & $\{06\}$ & $\{1\}$  & $\{1\}$ & $\{1\}$ \\
			  
			  && wash  (Delicates) & $0.39$ & $\{06\}$ & $\{1\}$  & $\{1\}$ & $\{1\}$ \\
			  && spin  (Delicates) & $0.33$ & $\{06\}$ & $\{1\}$  & $\{1\}$ & $\{1\}$ \\
			  && rinse (Delicates) & $0.01$ & $\{06\}$ & $\{1\}$  & $\{1\}$ & $\{1\}$ \\
\hline
\multirow{2}{*}{Clothes Dryer}  & \multirow{2}{*}{GE WSM2420D3WW}  
			& off 				& $0$ & $\{07\}$ & $\{0\}$ & $\{0\}$ & $\{0\}$ \\
		    && on (Regular)  	& $5.76$ & $\{07\}$ & $\{1\}$ & $\{1\}$ & $\{1\}$ \\    
		    && on (Perm-Press)  & $5.25$ & $\{07\}$ & $\{1\}$ & $\{1\}$ & $\{1\}$ \\    
		    && on (Timed)		& $5.95$ & $\{07\}$ & $\{1\}$ & $\{1\}$ & $\{1\}$ \\
\hline
\multirow{3}{*}{Oven}      & \multirow{3}{*}{Kenmore 790.91312013}  &    
			   off  	& $0$ & $\{05\}$ & $\{0\}$ & $\{0\}$ & $\{0\}$ \\
		     && bake	& $8.46$ & $\{05,\  01\}$ & $\{1, \  0.017^{\circ}C\}$ & $\{1, \  0.009^{\circ}C\}$ & $\{1, \  0.004^{\circ}C\}$ \\
		     && broil	& $10.56$	& $\{05,\  01\}$ & $\{1.25,\ 0.02^{\circ}C\}$  & $\{1.25,\ 0.01^{\circ}C\}$ & $\{1.25,\ 0.005^{\circ}C\}$ \\
\hline
\multirow{2}{*}{Dishwasher}   & \multirow{2}{*}{Kenmore 665.13242K900}  
			&  off  	& $0$    & $\{08\}$ & $\{0\}$ & $\{0\}$ & $\{0\}$ \\
		  	&& wash  	& $1.12$ & $\{08\}$ & $\{1\}$ & $\{1\}$ & $\{1\}$ \\
		  	&& rinse  	& $1.14$ & $\{08\}$ & $\{1\}$ & $\{1\}$ & $\{1\}$ \\
		  	&& dry  	& $0.63$ & $\{08\}$ & $\{1\}$ & $\{1\}$ & $\{1\}$ \\
\hline
\end{tabular}
}
\end{table}

\memo{switch columns state property with consumption }

\subsection{Physical models}
\label{sec:ds_pred_models}
\vspace{-4pt}

In this section we describe the physical models used to compute the 
effects values $\Delta$ of the actuators' actions on a predictive model 
(see Table \ref{tab:actuators}).
These values, in turn, are adopted within the SHDS predictive models as 
described in Equation \eqref{eq:sensorModel}.


\smallskip\noindent\textbf{Battery (Dis)charge Model}
The battery charge/discharge model we adopt for our battery-powered devices is as follows. 
For a given battery $b$ with capacity $Q_b$ (expressed in KWh), voltage $V_b$, 
and electric charge $E_b = \frac{V_b}{Q_b}$ (expressed in ampere-hour (Ah)), 
and assuming a 100\% charging/discharging  efficiency, the battery charge time 
$b_\alpha^+$ and discharge time $b_\alpha^-$ are computed respectively as: 
\begin{equation}
\begin{split}
b_\alpha^+ = \frac{E_b}{C^+}; \qquad
b_\alpha^- = \frac{E_b}{C^-},
\end{split}
\end{equation}
expressed in hours, where $C^+$ and $C^-$ are, respectively, the charging amperage and the in-use amperage.
Following \url{https://goo.gl/l5TGtz} and \url{https://goo.gl/NmO0fY} we report the 
battery parameters for our Electric Vehicle and robotic vacuum cleaner in Table \ref{tab:battery}.
The devices' action effects $\Delta$ for charging and discharging time are computed by 
dividing the total charging and discharging times by $|\setf{T}|$. 

\begin{table}[!tb]
\resizebox{0.6\textwidth}{!}
{
\caption{Electric vehicles \cite{telsaS} and robotic vacuum cleaner\cite{roomba} batteries physical model. \label{tab:battery}}
	\begin{tabular}{r r r r | r}
	\hline
		& \multicolumn{3}{c|}{Tesla Model S} & {iRobot Roomba 880} \\
		& Slow Charge & Regular Charge & Super Charger \\
	\hline
	$V_b$			&240	&240	 &240		&120\\
	$E_b$			&354 Ah & 354 Ah &354 Ah	& 3 Ah\\
	$C^+$			&48 A	& 72 A &500 A	 	& 1.25 A \\
	$C^-$			&60 A	& 60 A &60 A		& 0.75 A \\
	$b_\alpha^+$ 	&7 hr 22 min & 5 hr &43 min			& 2 hr 24 min \\
	$b_\alpha^-$ 	&6 hr & 6 hr & 6 hr				& 4 hr \\
	\hline
	\end{tabular}
}
\end{table}

\smallskip\noindent\textbf{Air Temperature Model}
The air temperature predictive model is computed following standard principle of 
heating and ventilation \cite{mitchell2012principles}  and described as follows. 
Let $G$ be the ventilation conductance: 
$
	G = \dot{V} \cdot \rho_a \cdot \bar{h}, 
$
where $\dot{V}$ is the volume flow rate, set to $100$, 
$\rho_a$ is the density of the air, set to 0.75, 
and $\bar{h}$ is the specific heat of the air, set to 0.24 following~\cite{mitchell2012principles}.
The house heat loss coefficient $h_{\text{loss}}$ is:
\begin{equation}
	h_{\text{loss}} = 
		U_{\text{walls}} \cdot A_\text{walls} 
		+ U_{\text{roof}} \cdot A_\text{roof} 
		+ U_{\text{windows}} \cdot A_\text{windows}
		+ G
\end{equation}
where $U_{\text{walls}}$, $U_{\text{roof}}$, and $U_{\text{windows}}$ are 
respectively, the  heat transfer coefficients for the walls, roof, and windows 
of the house, and $A_\text{walls}$, $A_\text{roof}$, and $A_\text{windows}$ are 
respectively the the areas for walls, roof, and windows. Their values are 
provided in Table \ref{tab:house}.
Let $T_A$ and $T_Z$ be the current and a target temperatures; the heating load $\dot{L}_h$ is given by:
\begin{equation}
	\dot{L}_h = h_{\text{loss}} |T_Z - T_A|
\end{equation}
The heating load defines the quantity of heat per unit time (in BTU) that must 
be supplied in a building to reach the target temperature $T_Z$, from the 
given temperature $T_A$.
Given the heating load $\dot{L}_h$ and the heater capacity $C$ of a heater/cooler,  
the time the device needs to run to reach the desired temperature is given by:
$
\frac{L_h}{C}
$.\\
Heating or cooling load is also effected by the outdoor and indoor temperature difference. 
Consider the example where $T_A$ = $12^{\circ}C$ and $T_Z$ = $22^{\circ}C$, and the outdoor temperature changes from $T_A$ to $T_N$ = $8^{\circ}C$. 
We can calculate the new load due to change in temperature by the following relationship given below:
\begin{equation}
	\dot{L}_n = \dot{L}_h \cdot \frac{|T_Z - T_N|}{|T_Z - T_A|}
\end{equation}
The above expression shows that an outdoor temperature drops of $4^{\circ}C$, causes the heating load to increase by a factor of 1.4 (w.r.t. the previous heating load $T_A$). In our model we need to compute the change in temperature per time step ($\Delta$). This can be done using the heat loss relationship: 
\begin{equation}
	 \Delta = \frac{h_{\text{loss}}}{m \cdot c_p}
\end{equation}
where $m$ is the mass of the air and $c_p$ is the specific heat of air. 
In our model, $m$ depends on volume flow rate of an air in the house, and $c_p$ = $1KJ/Kg\cdot K$.

\smallskip\noindent\textbf{Water Temperature Model}
The rise in the water temperature per unit of time ($\Delta$ value) is dependent 
on the difference in the water temperature flowing into the water heater and 
the amount of water flowing out of the water heater, as well as water usage We considered a gas-fired demand water heater(tankless). The water usage depends on household size and multiple user activities 
To calculate the water temperature, in our model, we used the highest potential peak water usage following 
\cite{waterHeater,waterUsage}, and 
corresponding to 26.50 liters/min (small house), 29.34 liters/min (medium house), and  38.38 liters/min (large house).
The rise in temperature is $39^{\circ}C$ for 18.93 liters/minute of water usage \cite{waterHeater}. Thus the rise in temperature for our small, medium, and large house, are, respectively, $27.9^{\circ}C$, $25.2^{\circ}C$, and $19.2^{\circ}C$.

\memo{Can you use liters rather than gallons?}

\smallskip\noindent\textbf{Cleanliness Model}
Our floor cleanliness model is computed by using the equation:
$
  T = \frac{A}{0.313}
$
where, $A$ represents the area of the room (in m$^2$) and 
$T$ is the amount of time (in minutes) it takes the robotic vacuum cleaner to vacuum the entire room. 
Consumer reports found that it took 57 minutes for the Roomba to clean a 17.84 m$^2$ room \cite{roomba} (which is approximately $0.313 m^2$/min).
In our experiment's datasets we use three different areas. $A_{small} = 48$, $A_{medium} = 96$, and $A_{large} = 180$. 
Thus the the estimated times to cover a 100\% floor for the small, medium, and large houses are, repsectively:  
$T=$ 153.35, 306.71, and 575.08 minutes.
The corresponding  $\Delta$ value of Table \ref{tab:actuators} (which is a percentage) is computed as: $\Delta = \frac{100\%}{T}$

\smallskip
All other predictive models (e.g., laundry wash and dry, bake, dish cleanliness, etc.) simply capture the time needed for a device to achieve the required goals by checking that accumulated device effects achieves the desired property. This is discussed in the dataset generation, in Section \ref{sec:experiments}.

\subsection{Scheduling Rules}
\label{sec:schduling_rules}
\vspace{-4pt}

We report, as follows, the complete Backus-Naur Form (BNF) for the \emph{scheduling rules} for a smart home $h_i \in \setf{H}$, introduced in Section \ref{sec:dev_schedules}
\begin{equation*} 
{\footnotesize
	\label{bnf}
\begin{split}
	\angles{rules} 
		&\defeq  \angles{simple rule} \ | \  \angles{simple rule} \land \angles{rules} \\
	\angles{simple rule} 
		&\defeq \angles{active rule} \ | \ \angles{passive rule} \\
	\angles{active rule} 
		&\defeq \angles{location} \angles{state property} \angles{relation} \angles{goal state} \angles{time} \\
	\angles{passive rule} 
		&\defeq \angles{location} \angles{state property} \angles{relation} \angles{goal state}\\
	\angles{location} 
		&\defeq \ell \in \setf{L}_i \\ 
	\angles{state property} 
		&\defeq s \in \setf{P}_H \ | \ s \in \setf{P}_Z \\ 
	\angles{relation} 
		&\defeq \ \leq \ | \ <  \ | \ =  \ | \ \neq \ | \ > \ | \ \geq \\
	\angles{goal state} 
		&\defeq \textit{sensor state} \ | \ \textit{actuator state} \\
	\angles{time} 
		&\defeq \text{at } \angles{T} \ | \ \text{before } \angles{T} \ | \ 
		\text{after } \angles{T} \ | \ \text{within } [\angles{T}, \angles{T}] \ | \ 
		 \text{for } \angles{T} \text{ time units}\\
	\angles{T} &\defeq t \in \setf{T}
\end{split}
}
\end{equation*}
In our dataset the device states are mapped to numeric values, i.e., $\Omega_p \!=\! \mathbb{N}$, for all $p \in \setf{P}_H \cup \setf{P}_Z$.

\subsection{Pricing Schema}
\label{sec:price_schema}
\vspace{-4pt}

For the evaluation of our SHDS datasets we adopted a pricing schema used by the Pacific Gas \& Electric Co. for its customers in parts of California,\footnote{\small \url{https://goo.gl/vOeNqj}} which accounts for 7 tiers ranging from \$0.198 per kWh to \$0.849 per kWh, reported in Table \ref{tab:pschema}

\begin{table}[!t]
\caption{Pacific Gas \& Electric Co.~pricing schema \label{tab:pschema}}
\begin{tabular}{l | c c c c c c}
\hline
time start & 0:00 & 8:00  & 12:00 & 14:00 & 18:00 & 22:00 \\
time end   & 7:59 & 11:59 & 13:59 & 17:59 & 21:59 & 23:59 \\
price (\$) & 0.198 		 & 0.225		& 0.249			& 0.849			& 0.225			& 0.198\\
\hline
\end{tabular}
\end{table}

\section{SHDS Dataset}
\label{sec:experiments}
\vspace{-4pt}

We now introduce a dataset for the SHDS problem for DCOPs.  We generate synthetic microgrid instances sampling neighborhoods in three cities in the United States (Des Moines, IA; Boston, MA; and San Francisco, CA) and estimate the density of houses in each city. The average density (in houses per square kilometers) is 718 in Des Moines, 1357 in Boston, and 3766 in San Francisco. For each city, we created a 200m$\times$200m grid, where the distance between intersections is 20m, and randomly placed houses in this grid until the density is the same as the sampled density. We then divided the city into $k$ (=$|\mathcal{H}|$) coalitions, where each home can communicate with all homes in its coalition. Finally, we ensure that there no two coalitions are disjoint. 
Tables \ref{tab:exp_DM}, \ref{tab:exp_BO}, and \ref{tab:exp_SF} report, respectively, the Des Moines, Boston, and San Francisco instances, where we vary (\textit{i}) the number of agents ($n$)---up to 1883 for the largest instances---, (\textit{ii}) the number of coalitions ($k$), and (\textit{iii}) the number of actuators within each home $m$. 

Each home device has an associated active scheduling rule that is randomly generated, and a number of passive rules that must always hold. The parameters to generate active and passive rules are reported, respectively, in Table \ref{tab:scheduling_rules_a} and Table \ref{tab:scheduling_rules_p}.
The time predicates for these rules are generated at random within the given horizon. Additionally, the relations $r$ and goals states $g_i$ are randomly generated sampling from the sets corresponding, respectively, to the 
columns $\angles{relation}$ and $\angles{goal state}$ of Table \ref{tab:scheduling_rules_a}.

We generate a total of $2351$ problem instances (available at: \url{https://github.com/persoon/SHDS_dataset}). We set $H=12$, and report in Table \ref{tab:ass} a summary of the parameters' settings for our smart homes physical models.

\begin{table}[!t]
\caption{Scheduling (active) rules  \label{tab:scheduling_rules_a}}
{
\begin{tabular}{c c c c c}
\hline
$\angles{location}$ & $\angles{state property}$ & $\angles{relation}$ & $\angles{goal state}$ & $\angles{time}$\\
\hline
\tinytt{Room} 	& \tinytt{air temperature} 		& $r \in \{ \leq, <, =, >, \geq \}$  & $g_1 \in [14, 28]$  & $\angles{time}$ \\
\tinytt{Room} 	& \tinytt{floor cleanliness} 	& $r \in \{ =, >, \geq \}$	& $g_2 \in [0, 100]$  & $\angles{time}$ \\
\tinytt{Electric Vehicle}  & \tinytt{charge}	& $r \in \{ =, >, \geq \}$ & $g_3 \in [0, 100]$  & $\angles{time}$\\
\tinytt{Water heater} & \tinytt{temperature} 	& $r \in \{ \leq, <, =, >, \geq \}$  & $g_4 \in [10, 45]$  & $\angles{time}$\\
\tinytt{Clothes Washer} & \tinytt{laundry wash} & $r \in \{ = \}$ & $g_5 \in \{45, 60\}$  & $\angles{time}$\\
\tinytt{Clothes Drier} 	& \tinytt{laundry dry} 	& $r \in \{ = \}$ & $g_6 \in \{45, 60\}$  & $\angles{time}$\\
\tinytt{Oven} 	& \tinytt{bake} 	            & $r \in \{ = \}$  & $g_7 \in \{30, 40, 60, 120\}$  & $\angles{time}$\\
\tinytt{Dishwasher} & \tinytt{dishes cleanliness} & $r \in \{ = \}$  & $g_8 \in \{45, 60\}$  & $\angles{time}$\\
\hline
\end{tabular}
}
\end{table}

\begin{table}[!tb]
\caption{Scheduling (active) rules  \label{tab:scheduling_rules_p}}
\resizebox{\textwidth}{!}
{
\begin{tabular}{c c c c c c c c c}
\hline
$\angles{location}$ & $\angles{state property}$ & $\angles{relation}$ & $\angles{goal state}$ 
&~~~~~& 
$\angles{location}$ & $\angles{state property}$ & $\angles{relation}$ & $\angles{goal state}$ \\
	\tinytt{Room} 	   & \tinytt{air temperature}   &  $\geq$		        & 0 
	&&
	\tinytt{EV} 	   & \tinytt{charge} 	     	&  $\leq$		        & 100 \\
	\tinytt{Room} 	   & \tinytt{air temperature}   &  $\leq$		        & 30
	&&
	\tinytt{Water heater} 	   & \tinytt{temperature}  	&  $\geq$		  & 10 \\
	\tinytt{Room} 	   & \tinytt{floor cleanliness}   &  $\geq$		        & 0
	&&
	\tinytt{Water heater} 	   & \tinytt{temperature} 	&  $\leq$		  & 42 \\
	\tinytt{Ooom} 	   & \tinytt{floor cleanliness}   &  $\leq$		        & 100
	&&
	\tinytt{Oven} 	   & \tinytt{bake} 		&  $\leq$		  & $g_7$ \\
	\tinytt{Roomba} 	   & \tinytt{charge}  		&  $\geq$		        & 0
	&&
	\tinytt{Clothes Washer} 	   & \tinytt{laundry wash} 	&  $\leq$		  & $g_6$ \\
	\tinytt{Roomba} 	   & \tinytt{charge} 	    	&  $\leq$		        & 100
	&&
	\tinytt{Clothes Drier} 	   & \tinytt{laundry dry} 	&  $\leq$		  & $g_7$ \\
	\tinytt{EV} 	   & \tinytt{charge}  		&  $\geq$		        & 0
	&&
	\tinytt{Dishwasher} & \tinytt{dishes cleanliness} 		&  $\leq$		  & $g_8$ \\
	\hline
\end{tabular}
}
\end{table}

Additionally, we provide upper bounds (in the \emph{obj} column) for each for the instances 
(Tables \ref{tab:exp_DM}, \ref{tab:exp_BO}, and \ref{tab:exp_SF} report a subset of the SHDS dataset) by solving an uncoordinated DCOP, where each agent reports its best schedule found with a local Constraint Programming solver\footnote{\small We adopt the JaCoP solver (\url{http://www.jacop.eu/})} as subroutine, within a 10 seconds timeout.

\begin{table}[!tb]
\caption{Physical models: Values and assumptions \label{tab:ass}}
\resizebox{0.8\textwidth}{!}
{
	\begin{tabular}{c c c c c}
	\hline
			Physical model  & Parameter 		& Value (small house) & Value (medium house) & Value (large house) \\
	\hline
	
	\multirow{6}{*}{Air Temperature}	&	$\dot{V}$		&  100 		& 200 		& 400 \\
										&	$m$				& 148.48	&296.86		&593.75	\\
										&	$c_p$			& 1.0		& 1.0		& 1.0	\\
										&	$\rho_a$		&  0.75		& 0.75 		& 0.75 	\\
										&	$\bar{h}$		&  0.24		&	0.24		& 0.24 \\
										&   $h\_{\text loss}$ & 352.24		& 544 		& 764.75 \\
										&	$T_Z$			& 22		& 22			& 22	\\				
										&	$T_A$			& 10		& 10			& 10	\\
										&   $\dot{L}_n$	    & 4226.88   & 6528          & 9177 \\
	\hline
	\multirow{3}{*}{Floor Cleanliness}	& $A$	& 48 m$^2$	& 96 m$^2$	& 180 m$^2$	\\
										& $T$	& 153.35 min	& 306.71 min	& 575.08 min\\
										& $\Delta$	& 0.652\%	& 0.326\%	& 0.174\%\\
	\hline
	\multirow{3}{*}{Water Temperature}	& household size	&2	&3		&4		\\
										& liters/min usage		&26.50	&29.34	&38.38	\\
										& $\Delta$		&27.9$^{\circ}C$ &25.2$^{\circ}C$	&19.2$^{\circ}C$	\\
	\hline
	
	\end{tabular}
}
\end{table}

\memo{What are the $\Delta T$ in Table 7?}

\newcolumntype{R}[1]{>{\raggedleft\let\newline\\\arraybackslash\hspace{0pt}}m{#1}}

\begin{table}[!h]
\resizebox{0.9\textwidth}{!}
{
\begin{tabular}{| l | r | r | r | r | R{1.75cm} | R{1.5cm} | R{1.5cm} | R{1.65cm} |}
\hline
instance & $n$ & $k$ & $m$ & sim.~time (s) & obj &avg price (\$) &avg power (kWh) &largest peak (kW) \\
\hline
\hline
dm\_7\_1\_3		&7	&1	&3	&28.00	 &458.27	 &1.11	&5.55	&8.75		\\
dm\_7\_1\_4		&7	&1	&4	&36.00	 &1385.70	 &1.83	&9.18	&16.85		\\
dm\_7\_1\_5		&7	&1	&5	&26.00	 &1936.87	 &2.36	&11.30	&20.21		\\
dm\_7\_1\_6		&7	&1	&6	&51.00	 &5122.10	 &4.06	&19.78	&29.55		\\
\hline
dm\_7\_4\_6		&7	&4	&6	&69.00	 &52335.95	 &4.08	&19.57	&32.02		\\
\hline
dm\_21\_1\_3	&21	&1	&3	&25.00	 &1280.79	 &1.06	&5.28	&26.28		\\
dm\_21\_1\_4	&21	&1	&4	&18.00	 &4101.46	 &1.80	&9.06	&48.62		\\
dm\_21\_1\_5	&21	&1	&5	&12.00	 &6070.22	 &2.44	&11.69	&59.66		\\
dm\_21\_1\_6	&21	&1	&6	&29.00	 &16252.47	 &4.06	&20.03	&92.30		\\
\hline
dm\_21\_16\_6	&21	&16	&6	&15.00	 &16435.40	 &40.43	&200.72	&944.80		\\
\hline
dm\_35\_1\_3	&35	&1	&3	&36.00	 &2349.93	 &1.12	&5.58	&44.52		\\
dm\_35\_1\_4	&35	&1	&4	&45.00	 &6982.77	 &1.81	&9.06	&85.63		\\
dm\_35\_1\_5	&35	&1	&5	&40.00	 &10055.69	 &2.48	&11.73	&99.33		\\
dm\_35\_1\_6	&35	&1	&6	&52.00	 &26355.29	 &4.02	&19.67	&154.89		\\
\hline
dm\_35\_32\_6	&35	&32	&6	&22.00	 &26215.24	 &4.03	&19.66	&152.40	\\
\hline
dm\_71\_1\_3	&71	&1	&3	&71.00	 &4814.35	 &1.13	&5.62	&91.13		\\
dm\_71\_1\_4	&71	&1	&4	&30.00	 &14548.58	 &1.83	&9.20	&170.78	\\
dm\_71\_1\_5	&71	&1	&5	&100.00	 &20180.40	 &2.43	&11.63	&201.66	\\
dm\_71\_1\_6	&71	&1	&6	&102.00	 &53406.99	 &4.03	&19.72	&314.30	\\
\hline
dm\_71\_64\_6	&71	&64	&6	&49.00	 &53170.13	 &4.01	&19.63	&316.57	\\
\hline
dm\_251\_1\_3	&251 &1	&3	&161.00	 &17167.95	 &1.13	&5.62	&320.5	\\
dm\_251\_1\_4	&251 &1	&4	&151.00	 &50798.96	 &1.82	&9.14	&599.9	\\
dm\_251\_1\_5	&251 &1	&5	&286.00	 &73057.18	 &2.48	&11.80	&721.0	\\
dm\_251\_1\_6	&251 &1	&6	&535.00	 &188736.13  &4.02	&19.63	&1117.5	\\
\hline
dm\_251\_128\_6	&251 &128&6	&66.00	 &188258.87 &4.03 &1.96 &1110.78	\\
\hline 
\end{tabular}
}
\caption{Des Moines \label{tab:exp_DM}}
\end{table}

\begin{table}[!h]
\resizebox{0.9\textwidth}{!}
{
\begin{tabular}{| l | r | r | r | r | R{1.75cm} | R{1.5cm} | R{1.5cm} | R{1.65cm} |}
\hline
instance & $n$ & $k$ & $m$ & sim.~time (s) & obj &avg price (\$) &avg power (kWh) &largest peak (kW) \\
\hline
\hline
bo\_13\_1\_3	&13	&1	&3	&16.00	 &843.75	 &1.06	 &5.33	  &17.36 \\
bo\_13\_1\_4	&13	&1	&4	&32.00	 &2761.91	 &1.87	 &9.40	  &31.75 \\
bo\_13\_1\_5	&13	&1	&5	&80.00	 &3707.74	 &2.52	 &11.86  &37.01 \\
bo\_13\_1\_6	&13	&1	&6	&77.00	 &9181.61	 &4.03	 &19.41  &53.59 \\
\hline 
bo\_13\_8\_6	&13	&8	&6	&25.00	 &9513.77	 &4.02	 &19.50  &57.03 \\
\hline 
bo\_40\_1\_3	&40	&1	&3	&20.00	 &2651.70	 &1.11	 &5.53	  &50.03 \\
bo\_40\_1\_4	&40	&1	&4	&26.00	 &8016.06	 &1.84	 &9.23	  &92.99 \\
bo\_40\_1\_5	&40	&1	&5	&48.00	 &11335.13	 &2.50	 &11.75 &111.41 \\
bo\_40\_1\_6	&40	&1	&6	&63.00	 &30507.66	 &4.07	 &19.78 &179.32 \\
\hline 
bo\_40\_32\_6	&40	&32	&6	&29.00	 &29978.17	 &4.08	 &19.71 &177.39 \\
\hline 
bo\_67\_1\_3	&67	&1	&3	&5.00	 &4625.82	 &1.12	 &5.59	  &88.89 \\
bo\_67\_1\_4	&67	&1	&4	&25.00	 &13218.52	 &1.80	 &9.01	 &158.50 \\
bo\_67\_1\_5	&67	&1	&5	&26.00	 &19166.54	 &2.40	 &11.68 &193.20 \\
bo\_67\_1\_6	&67	&1	&6	&128.00	 &49565.08	 &4.00	&19.56	 &293.76 \\
\hline 
bo\_67\_64\_6	&67	&64	&6	&21.00	 &50847.42	 &4.04	 &19.70 &298.76 \\
\hline 
bo\_135\_1\_3	&135	&1	&3	&45.00	 &8981.98	 &1.11	 &5.53	 &172.25 \\
bo\_135\_1\_4	&135	&1	&4	&170.00	 &27602.08	 &1.83	 &9.16	 &325.76 \\
bo\_135\_1\_5	&135	&1	&5	&114.00	 &38668.73	 &2.45	 &11.64 &386.87 \\
bo\_135\_1\_6	&135	&1	&6	&184.00	 &102105.12 &4.04	 &19.77 &598.08 \\
\hline 
bo\_135\_128\_6	&135	&128&6	&78.00	 &101369.93 &4.04	 &19.69 &596.32 \\
\hline 
bo\_474\_1\_3	&474	&1	&3	&949.00	 &31864.64	 &1.11	 &5.56	 &600.63 \\
bo\_474\_1\_4	&474	&1	&4	&159.00	 &95450.24	 &1.82	 &9.11	 &1136.34\\
bo\_474\_1\_5	&474	&1	&5	&1230.00 &136395.09 &2.46	 &11.73 &1358.24\\
bo\_474\_1\_6	&474	&1	&6	&411.00	 &356417.23 &4.04	 &19.70 &2106.18\\
\hline
bo\_474\_256\_6	&474	&256&6	&49.40	 &35510.81 &4.03	 &19.66 &2092.18\\
\hline 
\end{tabular}
}
\caption{Boston \label{tab:exp_BO}}
\end{table}

\begin{table}[!h]
\resizebox{0.9\textwidth}{!}
{
\begin{tabular}{| l | r | r | r | r | R{1.75cm} | R{1.5cm} | R{1.5cm} | R{1.65cm} |}
\hline
instance & $n$ & $k$ & $m$ & sim.~time (s) & obj &avg price (\$) &avg power (kWh) &largest peak (kW) \\
\hline
\hline
sf\_37\_1\_3	&37		&1	&3	&121.00	 &2621.67	&1.13	&5.67 	&4.77 	\\
sf\_37\_1\_4	&37		&1	&4	&121.00	 &7666.59	&1.85 	&9.30 	&89.67		\\
sf\_37\_1\_5	&37		&1	&5	&151.00	 &10345.77	&2.53 	&11.67	&102.47 	\\
sf\_37\_1\_6	&37		&1	&6	&116.00	 &28216.43	&4.00 	&19.75	&166.31 	\\
\hline
sf\_37\_32\_6	&37		&32	&6	&30.00	 &26926.71	&4.00	&19.45	&158.23 	\\
\hline
sf\_112\_1\_3	&112	&1	&3	&31.00	 &75331.60	&1.11	&5.54 	&143.33 	\\
sf\_112\_1\_4	&112	&1	&4	&47.00	 &23099.46	&1.84	&9.25 	&270.51 	\\
sf\_112\_1\_5	&112	&1	&5	&244.00	 &32254.58	&2.44	&11.73	&319.10 	\\
sf\_112\_1\_6	&112	&1	&6	&59.00	 &84245.82	&4.04	&19.66	&493.40 	\\
\hline
sf\_112\_64\_6	&112	&64	&6	&78.00	 &85088.67	&4.06	&19.79	&496.49 	\\
\hline
sf\_188\_1\_3	&188	&1	&3	&162.00	 &124835.94	&1.11	&5.52 	&234.68 	\\
sf\_188\_1\_4	&188	&1	&4	&163.00	 &381943.73	&1.82	&9.14	&452.78 	\\
sf\_188\_1\_5	&188	&1	&5	&142.00	 &547287.48	&2.47	&11.82	&533.91 	\\
sf\_188\_1\_6	&188	&1	&6	&230.00	 &140592.96 &4.04	&19.62	&830.50 	\\
\hline
sf\_188\_128\_6	&188   &128	&6	&430.00	 &141294.31 &4.02	&19.68	&830.99 	\\
\hline
sf\_376\_1\_4	&376	&1	&4	&556.00	 &74580.82	&1.81	&9.06	&894.12 	\\
sf\_376\_1\_5	&376	&1	&5	&735.00	 &108104.93 &2.48	&11.74	&1072.98	\\
sf\_376\_1\_6	&376	&1	&6	&1033.00 &283038.14 &4.03	&19.69	&1667.89	\\
\hline
sf\_376\_256\_6	&376 &256   &6	&302.00	 &28306328 & 4.04	&19.73 &1652.88	\\
\hline 
\end{tabular}
}
\caption{San Francisco \label{tab:exp_SF}}
\end{table}

\section{Conclusions}
\vspace{-4pt}

\noindent With the proliferation of smart devices, the automation of smart home scheduling can be a powerful tool for demand-side management within the smart grid vision. 
In this paper we proposed the \emph{Smart Home Device Scheduling (SHDS)} problem, which formalizes the device scheduling and coordination problem across multiple smart homes as a multi-agent system, and its mapping to a DCOP. 
Furthermore, we described in great details the physical models adopted to model the smart home's sensors and actuators, as well as the physical model regulating the effect of the devices actions on the house environments properties (e.g., temperature, cleanliness).
Finally, we reported a realistic dataset for the SHDS problem for DCOPs which includes 2351 instances of increasing difficulty.

%
\footnotesize
\bibliographystyle{abbrv}
\bibliography{si,Fioretto,mas}

\begin{thebibliography}{10}

\bibitem{roomba}
Roomba 880 specs.
\newblock
  http://www.consumerreports.org/products/robotic-vacuum/roomba-880-290102/specs/.
\newblock [Online; accessed 18-February-2017].

\bibitem{waterHeater}
Sizing a new water heater.
\newblock \url{https://www.energy.gov/energysaver/sizing-new-water-heater}.
\newblock [Online; accessed 18-February-2017].

\bibitem{telsaS}
Tesla model s specifics.
\newblock https://www.tesla.com/models.

\bibitem{waterUsage}
Typical water used in normal home activities.
\newblock \url{http://www.pittsfield-mi.gov/DocumentCenter/View/285}.
\newblock [Online; accessed 18-February-2017].

\bibitem{farinelli:08}
A.~Farinelli, A.~Rogers, A.~Petcu, and N.~Jennings.
\newblock Decentralised coordination of low-power embedded devices using the
  {Max-Sum} algorithm.
\newblock In {\em AAMAS}, pages 639--646, 2008.

\bibitem{fioretto:cp16}
F.~Fioretto, W.~Yeoh, and E.~Pontelli.
\newblock A dynamic programming-based {MCMC} framework for solving {DCOP}s with
  {GPU}s.
\newblock In {\em Proceedings of the International Conference on Principles and
  Practice of Constraint Programming (CP)}, pages 813--831, 2016.

\bibitem{fioretto:aamas-17b}
F.~Fioretto, W.~Yeoh, and E.~Pontelli.
\newblock A multiagent system approach to scheduling devices in smart homes.
\newblock In {\em AAMAS}, page (to appear), 2017.

\bibitem{gershman:09}
A.~Gershman, A.~Meisels, and R.~Zivan.
\newblock {Asynchronous Forward-Bounding} for distributed {COP}s.
\newblock {\em JAIR}, 34:61--88, 2009.

\bibitem{hirayama:97}
K.~Hirayama and M.~Yokoo.
\newblock Distributed partial constraint satisfaction problem.
\newblock In {\em CP}, pages 222--236, 1997.

\bibitem{mailler:04}
R.~Mailler and V.~Lesser.
\newblock Solving distributed constraint optimization problems using
  cooperative mediation.
\newblock In {\em AAMAS}, pages 438--445, 2004.

\bibitem{modi:05}
P.~Modi, W.-M. Shen, M.~Tambe, and M.~Yokoo.
\newblock {ADOPT}: Asynchronous distributed constraint optimization with
  quality guarantees.
\newblock {\em AIJ}, 161(1--2):149--180, 2005.

\bibitem{nguyen:13}
D.~T. Nguyen, W.~Yeoh, and H.~C. Lau.
\newblock Distributed {G}ibbs: A memory-bounded sampling-based {DCOP}
  algorithm.
\newblock In {\em AAMAS}, pages 167--174, 2013.

\bibitem{ottens:12}
B.~Ottens, C.~Dimitrakakis, and B.~Faltings.
\newblock {DUCT}: An upper confidence bound approach to distributed constraint
  optimization problems.
\newblock In {\em AAAI}, pages 528--534, 2012.

\bibitem{pearce:07}
J.~Pearce and M.~Tambe.
\newblock Quality guarantees on k-optimal solutions for distributed constraint
  optimization problems.
\newblock In {\em IJCAI}, pages 1446--1451, 2007.

\bibitem{petcu:05}
A.~Petcu and B.~Faltings.
\newblock A scalable method for multiagent constraint optimization.
\newblock In {\em IJCAI}, pages 1413--1420, 2005.

\bibitem{petcu:07b}
A.~Petcu, B.~Faltings, and R.~Mailler.
\newblock {PC-DPOP}: A new partial centralization algorithm for distributed
  optimization.
\newblock In {\em IJCAI}, pages 167--172, 2007.

\bibitem{mitchell2012principles}
M.~J. W and J.~E. ~, Braun.
\newblock {\em Principles of Heating, Ventilation and Air Conditioning in
  Buildingss}.
\newblock Wiley, 2012.

\bibitem{yeoh:12}
W.~Yeoh and M.~Yokoo.
\newblock Distributed problem solving.
\newblock {\em AI Magazine}, 33(3):53--65, 2012.

\bibitem{zhang:05}
W.~Zhang, G.~Wang, Z.~Xing, and L.~Wittenberg.
\newblock Distributed stochastic search and distributed breakout: Properties,
  comparison and applications to constraint optimization problems in sensor
  networks.
\newblock {\em AIJ}, 161(1--2):55--87, 2005.

\end{thebibliography}
\vspace{-4pt}  

\end{document}